\documentclass[letterpaper]{article} % DO NOT CHANGE THIS
\usepackage{aaai25}  % Enable for the camera ready version
\usepackage{times}  % DO NOT CHANGE THIS
\usepackage{helvet}  % DO NOT CHANGE THIS
\usepackage{courier}  % DO NOT CHANGE THIS
\usepackage[hyphens]{url}  % DO NOT CHANGE THIS
\usepackage{graphicx} % DO NOT CHANGE THIS
\usepackage{natbib}  % DO NOT CHANGE THIS AND DO NOT ADD ANY OPTIONS TO IT
\usepackage{caption} % DO NOT CHANGE THIS AND DO NOT ADD ANY OPTIONS TO IT
\usepackage{algorithm}
\usepackage{algorithmic}

\usepackage{newfloat}
\usepackage{listings}
\DeclareCaptionStyle{ruled}{labelfont=normalfont,labelsep=colon,strut=off} % DO NOT CHANGE THIS
\floatstyle{ruled}
\newfloat{listing}{tb}{lst}{}
\floatname{listing}{Listing}
\PassOptionsToPackage{numbers, compress}{natbib}

\usepackage{amsmath}
\usepackage{multirow}
\usepackage{xcolor}
\usepackage{booktabs} % For better table lines
\usepackage{url}
\title{Docling: An Efficient Open-Source Toolkit for AI-driven Document Conversion}
\author{
    Nikolaos~Livathinos \equalcontrib,
    Christoph~Auer \equalcontrib,
    Maksym~Lysak,
    Ahmed~Nassar,
    Michele~Dolfi,
    Panagiotis~Vagenas,
    Cesar~Berrospi,
    Matteo~Omenetti,
    Kasper~Dinkla,
    Yusik~Kim,
    Shubham~Gupta,
    Rafael~Teixeira~de~Lima,
    Valery~Weber,
    Lucas~Morin,
    Ingmar~Meijer,
    Viktor~Kuropiatnyk,
    Peter~W.~J.~Staar
}
\affiliations{
    IBM Research, R\"uschlikon, Switzerland \\
    Please send correspondence to: \texttt{deepsearch-core@zurich.ibm.com} \\
}

\begin{document}

\maketitle

% \begin{figure}
% \centering
% \includegraphics[width=0.25\linewidth]{figures/docling.jpeg}
% % \caption{\label{fig:docling}Docling - easy PDF processing.}
% \end{figure}

\begin{abstract}
% We introduce \textit{Docling}, an easy to use, self-contained, MIT-licensed open-source package for document conversion. It is powered by state-of-the-art specialized AI models for layout analysis (DocLayNet) and table structure recognition (TableFormer), and runs efficiently on commodity hardware in a small resource budget. The code interface allows for easy extensibility and addition of new features and models. The open-source community is fully engaged in using, promoting and developing Docling, which crashed already the milestone of 10k stars on GitHub and was reported the nr 1 project in the GitHub worldwide trending repositories.

We introduce \textit{Docling}, an easy-to-use, self-contained, MIT-licensed, open-source toolkit for document conversion, that can parse several types of popular document formats into a unified, richly structured representation. It is powered by state-of-the-art specialized AI models for layout analysis (DocLayNet) and table structure recognition (TableFormer), and runs efficiently on commodity hardware in a small resource budget. Docling is released as a Python package and can be used as a Python API or as a CLI tool. Docling's modular architecture and efficient document representation %, known as \textit{DoclingDocument}, 
make it easy to implement extensions, new features, models, and customizations. Docling has been already integrated in other popular open-source frameworks (e.g., LangChain, LlamaIndex, spaCy), making it a natural fit for the processing of documents and the development of high-end applications.
The open-source community has fully engaged in using, promoting, and developing for Docling, which gathered 10k stars on GitHub in less than a month and was reported as the No. 1 trending repository in GitHub worldwide in November 2024.
\end{abstract}

\begin{links}
    \link{Repository}{https://github.com/DS4SD/docling}
\end{links}

\section{Introduction}

Converting documents back into a unified machine-processable format has been a major challenge for decades due to their huge variability in formats, weak standardization and printing-optimized characteristic, which often discards structural features and metadata. With the advent of LLMs and popular application patterns such as retrieval-augmented generation (RAG), leveraging the rich content embedded in PDFs, Office documents, and scanned document images has become ever more relevant.
In the past decade, several powerful document understanding solutions have emerged on the market, most of which are commercial software, SaaS offerings on hyperscalers~\cite{auer2022delivering} and most recently, multimodal vision-language models. Typically, they incur a cost (e.g., for licensing or LLM inference) and cannot be run easily on local hardware. 
Meanwhile, only a handful of different open-source tools cover PDF, MS Word, MS PowerPoint, Images, or HTML conversion, leaving a significant feature and quality gap to proprietary solutions.
%such as OpenAI's GPT4o, Claude 3.5 Sonnet or Google Gemini. 

With \textit{Docling}, we recently open-sourced a very capable and efficient document conversion tool which builds on the powerful, specialized AI models and datasets for layout analysis and table structure recognition that we developed and presented in the recent past~\cite{pagemodel, dln, lysak_optimized_2023}. Docling is designed as a simple, self-contained Python library with permissive MIT license, running entirely locally on commodity hardware. Its code architecture allows for easy extensibility and addition of new features and models. 
Since its launch in July 2024, Docling has attracted considerable attention in the AI developer community and ranks top on GitHub's monthly trending repositories with more than 10,000 stars at the time of writing. On October 16, 2024, Docling reached a major milestone with version 2, introducing several new features and concepts, which we outline in this updated technical report, along with details on its architecture, conversion speed benchmarks, and comparisons to other open-source assets.

The following list summarizes the features currently available on Docling:

\begin{itemize}
    \item Parses common document formats (PDF, Images, MS Office formats, HTML) and exports to Markdown, JSON, and HTML.
    \item Applies advanced AI for document understanding, including detailed page layout, OCR, reading order, figure extraction, and table structure recognition.
    \item Establishes a unified \texttt{DoclingDocument} data model for rich document representation and operations.
    \item Provides fully local execution capabilities making it suitable for sensitive data and air-gapped environments.
    \item Has an ecosystem of plug-and-play integrations with prominent generative AI development frameworks, including LangChain and LlamaIndex.
    % \item Offers a simple command-line interface.
    \item Can leverage hardware accelerators such as GPUs.
\end{itemize}

\begin{figure*}[ht]
    \centering
    \includegraphics[width=0.85\linewidth]{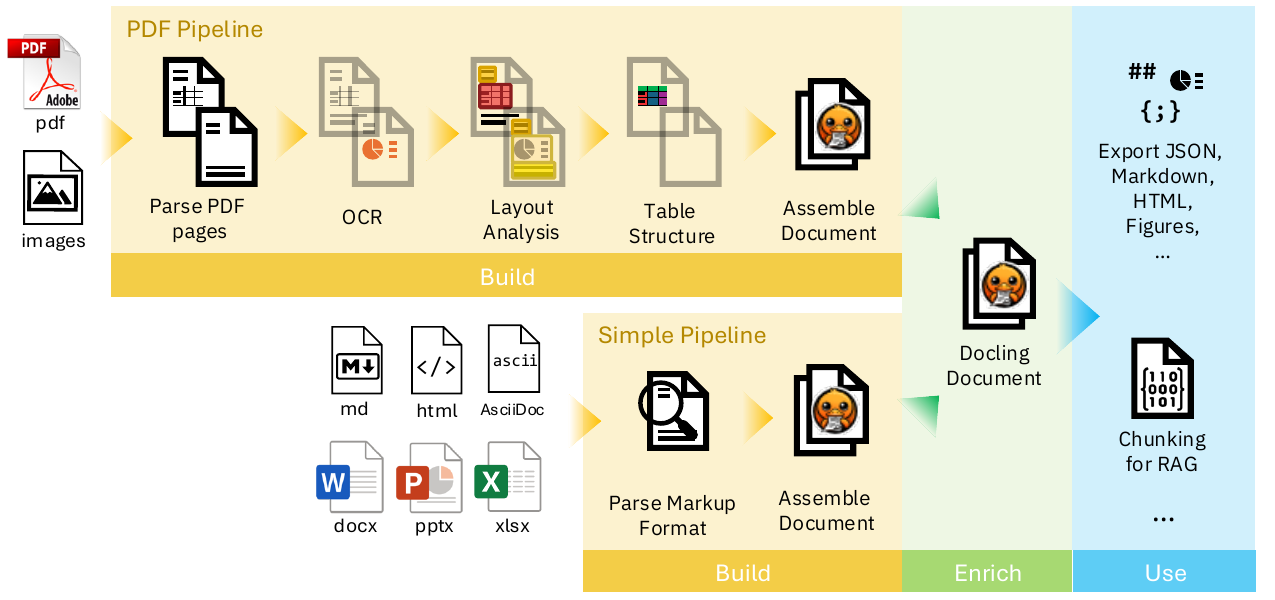}
    \caption{Sketch of Docling's pipelines and usage model. Both PDF pipeline and simple pipeline build up a \texttt{DoclingDocument} representation, which can be further enriched. Downstream applications can utilize Docling's API to inspect, export, or chunk the document for various purposes.}
    \label{fig:arch}
\end{figure*}

\section{State of the Art}

Document conversion is a well-established field with numerous solutions already available on the market. These solutions can be categorized along several key dimensions, including open vs. closed source, permissive vs. restrictive licensing, Web APIs vs. local code deployment, susceptibility to hallucinations, conversion quality, time-to-solution, and compute resource requirements.

The most popular conversion tools today leverage vision-language models (VLMs), which process page images to text and markup directly. Among proprietary solutions, prominent examples include GPT-4o (OpenAI), Claude (Anthropic), and Gemini (Google). In the open-source domain, LLaVA-based models, such as LLaVA-next, are noteworthy. However, all generative AI-based models face two significant challenges. First, they are prone to hallucinations, i.e., their output may contain false information which is not present in the source document — a critical issue when faithful transcription of document content is required. Second, these models demand substantial computational resources, making the conversion process expensive. Consequently, VLM-based tools are typically offered as SaaS, with compute-intensive operations performed remotely in the cloud.

A second category of solutions prioritizes on-premises deployment, either as Web APIs or as libraries. Examples include Adobe Acrobat, Grobid, Marker, MinerU, Unstructured, and others. These solutions often rely on multiple specialized models, such as OCR, layout analysis, and table recognition models. Docling falls into this category, leveraging modular, task-specific models which recover document structures and features only. All text content is taken from the programmatic PDF or transcribed through OCR methods. This design ensures faithful conversion, without the risk of generating false content. However, it necessitates maintaining a diverse set of models for different document components, such as formulas or figures.

Within this category, Docling distinguishes itself through its permissive MIT license, allowing organizations to integrate Docling into their solutions without incurring licensing fees or adopting restrictive licenses (e.g., GPL). Additionally, Docling offers highly accurate, resource-efficient, and fast models, making it well-suited for integration with many standard frameworks.

In summary, Docling stands out as a cost-effective, accurate and transparent open-source library with a permissive license, offering a reliable and flexible solution for document conversion.

\section{Design and Architecture}

Docling is designed in a modular fashion with extensibility in mind, and it builds on three main concepts: pipelines, parser backends, and the \texttt{DoclingDocument} data model as its centerpiece (see Figure~\ref{fig:arch}). Pipelines and parser backends share the responsibility of constructing and enriching a \texttt{DoclingDocument} representation from any supported input format. The \texttt{DoclingDocument} data model with its APIs enable inspection, export, and downstream processing for various applications, such as RAG.

\subsection{Docling Document}

Docling v2 introduces a unified document representation, \texttt{DoclingDocument}, as a Pydantic data model that can express various common document features, such as:

\begin{itemize}
    \item Text, Tables, Pictures, Captions, Lists, and more.
    \item Document hierarchy with sections and groups.
    \item Disambiguation between main body and headers, footers (furniture).
    \item Layout information (i.e., bounding boxes) for all items, if available.
    \item Provenance information (i.e., page numbers, document origin).
\end{itemize}

With this data model, Docling enables representing document content in a unified manner, i.e., regardless of the source document format.
%This representation can be losslessly serialized to JSON, offering a practical and universal means of storing and loading its contents.
% The \texttt{DoclingDocument} serializable to JSON, which offers a lossless way of storing and loading its contents.

Besides specifying the data model, the \texttt{DoclingDocument} class defines APIs encompassing document construction, inspection, and export. Using the respective methods, users can incrementally build a \texttt{DoclingDocument}, traverse its contents in reading order, or export to commonly used formats. Docling supports lossless serialization to (and deserialization from) JSON, and lossy export formats such as Markdown and HTML, which, unlike JSON, cannot retain all available meta information.

A \texttt{DoclingDocument} can additionally be passed to a chunker class, an abstraction that returns a stream of chunks, each of which captures some part of the document as a string accompanied by respective metadata. To enable both flexibility for downstream applications and out-of-the-box utility, Docling defines a chunker class hierarchy, providing a base type as well as specific subclasses.
%For instance, one such specific implementation primarily creates chunks aligned with document units such as paragraphs, but also leverages heuristics such as merging list items.
By using the base chunker type, downstream applications can leverage popular frameworks like LangChain or LlamaIndex, which provide a high degree of flexibility in the chunking approach. Users can therefore plug in any built-in, self-defined, or third-party chunker implementation.

% A typical chunking application is within RAG, whereby the string is used as an input to an embedding model, and the metadata may have various purposes such as filtering. When chunking the \texttt{DoclingDocument} object itself (and not a lossy export format like Markdown), the rich metadata available can also be leveraged for providing precise document-native grounding information like the page number and the bounding box of the context supporting the RAG answer, as currently done in the LlamaIndex \texttt{DoclingNodeParser} plugin.

\subsection{Parser Backends}

Document formats can be broadly categorized into two types:
\begin{enumerate}
    \item \textbf{Low-level formats}, like PDF files or scanned images. These formats primarily encode the visual representation of the document, containing instructions for rendering text cells and lines or defining image pixels. Most semantics of the represented content are typically lost and need to be recovered through specialized AI methods, such as OCR, layout analysis, or table structure recognition.
    \item \textbf{Markup-based formats}, including MS Office, HTML, Markdown, and others. These formats preserve the semantics of the content (e.g., sections, lists, tables, and figures) and are comparatively inexpensive to parse. 
\end{enumerate}

Docling implements several parser backends to read and interpret different formats and it routes their output to a fitting processing pipeline.
For PDFs Docling provides backends which: a) retrieve all text content and their geometric properties, b) render the visual representation of each page as it would appear in a PDF viewer.
For markup-based formats, the respective backends carry the responsibility of creating a \texttt{DoclingDocument} representation directly. 
For some formats, such as PowerPoint slides, element locations and page provenance are available, whereas in other formats (for example, MS Word or HTML), this information is unknown unless rendered in a Word viewer or a browser. The \textit{DoclingDocument} data model handles both cases.

\subsubsection*{PDF Backends}

% While several open-source PDF parsing python libraries are available, we faced major obstacles with all of them for multiple reasons, among which were restrictive licensing (e.g., pymupdf~\cite{pymupdf}), poor speed, or unrecoverable quality issues, such as merged text cells across far-apart text tokens or table columns (pypdfium, PyPDF)~\cite{pypdfium2,pypdf}.
While several open-source PDF parsing Python libraries are available, in practice we ran into various limitations, among which are restrictive licensing (e.g., pymupdf~\cite{pymupdf}), poor speed, or unrecoverable quality issues, such as merged text cells across far-apart text tokens or table columns (pypdfium, PyPDF)~\cite{pypdfium2,pypdf}.

We therefore developed a custom-built PDF parser, which is based on the low-level library \textit{qpdf}~\cite{qpdf}. Our PDF parser is made available in a separate package named \textit{docling-parse} and acts as the default PDF backend in Docling. As an alternative, we provide a PDF backend relying on \textit{pypdfium}~\cite{pypdfium2}.

\subsubsection*{Other Backends}

Markup-based formats like HTML, Markdown, or Microsoft Office (Word, PowerPoint, Excel) as well as plain formats like AsciiDoc can be transformed directly to a \texttt{DoclingDocument} representation with the help of several third-party format parsing libraries. For HTML documents we utilize \textit{BeautifulSoup}~\cite{beautifulsoup}, for Markdown we use the \textit{Marko} library~\cite{marko}, and for Office XML-based formats (Word, PowerPoint, Excel) we implement custom extensions on top of the \textit{python-docx}~\cite{python_docx}, \textit{python-pptx}~\cite{python_pptx}, and \textit{openpyxl}~\cite{openpyxl} libraries, respectively. During parsing, we identify and extract common document elements (e.g., title, headings, paragraphs, tables, lists, figures, and code) and reflect the correct hierarchy level if possible. 

\subsection{Pipelines}

Pipelines in Docling serve as an orchestration layer which iterates through documents, gathers the extracted data from a parser backend, and applies a chain of models to: a) build up the \texttt{DoclingDocument} representation and b) enrich this representation further (e.g., classify images).

Docling provides two standard pipelines. The \textit{StandardPdfPipeline} leverages several state-of-the-art AI models to reconstruct a high-quality \textit{DoclingDocument} representation from PDF or image input, as described in section~\ref{sec:pdf_pipeline}. The \textit{SimplePipeline} handles all markup-based formats (Office, HTML, AsciiDoc) and may apply further enrichment models as well.

Pipelines can be fully customized by sub-classing from an abstract base class or cloning the default model pipeline. This effectively allows to fully customize the chain of models, add or replace models, and introduce additional pipeline configuration parameters. To create and use a custom model pipeline, you can provide a custom pipeline class as an argument to the main document conversion API. % To use a custom model pipeline, the custom pipeline class to instantiate can be provided as an argument to the main document conversion API.

\section{PDF Conversion Pipeline\label{sec:pdf_pipeline}} % better name??

The capability to recover detailed structure and content from PDF and image files is one of Docling's defining features. In this section, we outline the underlying methods and models that drive the system.

Each document is first parsed by a PDF backend, which retrieves the programmatic text tokens, consisting of string content and its coordinates on the page, and also renders a bitmap image of each page to support downstream operations. Any image format input is wrapped in a PDF container on the fly, and proceeds through the pipeline as a scanned PDF document. 
Then, the standard PDF pipeline applies a sequence of AI models independently on every page of the document to extract features and content, such as layout and table structures. Finally, the results from all pages are aggregated and passed through a post-processing stage, which eventually assembles the \texttt{DoclingDocument} representation.

\subsection{AI Models}

As part of Docling, we release two highly capable AI models to the open-source community, which have been developed and published recently by our team. The first model is a layout analysis model, an accurate object detector for page elements~\cite{dln}. The second model is TableFormer~\cite{tableformer,lysak_optimized_2023}, a state-of-the-art table structure recognition model. We provide the pre-trained weights (hosted on Hugging Face) and a separate Python package for the inference code (\textit{docling-ibm-models}).

\subsubsection*{Layout Analysis Model}
Our layout analysis model is an object detector which predicts the bounding-boxes and classes of various elements on the image of a given page. Its architecture is derived from RT-DETR~\cite{lv2023detrs} and re-trained on DocLayNet~\cite{dln}, our popular human-annotated dataset for document-layout analysis, among other proprietary datasets. For inference, our implementation relies on the \textit{Hugging Face transformers}~\cite{wolf2020huggingfacestransformersstateoftheartnatural} library and the \textit{Safetensors} file format. All predicted bounding-box proposals for document elements are post-processed to remove overlapping proposals based on confidence and size, and then intersected with the text tokens in the PDF to group them into meaningful and complete units such as paragraphs, section titles, list items, captions, figures, or tables.

\subsubsection*{Table Structure Recognition}
The TableFormer model~\cite{tableformer}, first published in 2022 and since refined with a custom structure token language~\cite{lysak_optimized_2023}, is a vision-transformer model for table structure recovery. It can predict the logical row and column structure of a given table based on an input image, and determine which table cells belong to column headers, row headers or the table body. Compared to earlier approaches, TableFormer handles many characteristics of tables like partial or no borderlines, empty cells, rows or columns, cell spans and hierarchy on both column-heading and row-heading level, tables with inconsistent indentation or alignment and other complexities. For inference, our implementation relies on \textit{PyTorch}~\cite{pytorch2}. The PDF pipeline feeds all table objects detected in the layout analysis to the TableFormer model, by providing an image-crop of the table and the included text cells. TableFormer structure predictions are matched back to the PDF cells during a post-processing step, to avoid expensive re-transcription of the table image-crop, which also makes the TableFormer model language agnostic.

%We include two different configurations of the model:
%\begin{itemize}
%    \item Standard - includes more encoder / decoder layers, state of the art model %for table structure recognition. Handy for getting the best prediction quality, at the %cost of prediction time.
%    \item Fast - smaller configuration of the same architecture, that slightly %sacrifices potential quality of prediction over significant gains in speed. Very %useful when processing very large document corpus.
%\end{itemize}
%Both configurations are trained on the same set of public datasets: PubTabNet %~\cite{pubtabnet}, FinTabNet ~\cite{ftn}, PubTables1M  ~\cite{pub1m}, WikiTables (EN)  %~\cite{icpram23}, and fine-tuned by in-house annotated datasets for table header %understanding.

\subsubsection*{OCR}

% Docling provides support for OCR, for example to cover scanned PDFs or content in bitmaps images embedded on a page.
Docling utilizes OCR to convert scanned PDFs and extract content from bitmaps images embedded in a page. Currently, we provide integration with \textit{EasyOCR}~\cite{easyocr}, a popular third-party OCR library with support for many languages, and Tesseract as a widely available alternative. While EasyOCR delivers reasonable transcription quality, we observe that it runs fairly slow on CPU (see section~\ref{sec:perf}), making it the biggest compute expense in the pipeline. 

\subsubsection*{Assembly}

In the final pipeline stage, Docling assembles all prediction results produced on each page into the \textit{DoclingDocument} representation, as defined in the auxiliary Python package \textit{docling-core}. The generated document object is passed through a post-processing model which leverages several algorithms to augment features, such as correcting the reading order or matching figures with captions.

\section{Performance\label{sec:perf}}

In this section, we characterize the conversion speed of PDF documents with Docling in a given resource budget for different scenarios and establish reference numbers. 

Further, we compare the conversion speed to three popular contenders in the open-source space, namely \textit{unstructured.io}~\cite{unstructured_io}, \textit{Marker}~\cite{marker}, and \textit{MinerU}~\cite{wang2024mineruopensourcesolutionprecise}. All aforementioned solutions can universally convert PDF documents to Markdown or similar representations and offer a library-style interface to run the document processing entirely locally. We exclude SaaS offerings and remote services for document conversion from this comparison, since the latter do not provide any possibility to control the system resources they run on, rendering any speed comparison invalid.

%\textit{TODO: Check if we can even do this...}
%Also, we provide anecdotal evidence of typical quality characteristics seen from all tested assets. While we previously published accuracy metrics for the individual models in Docling~\cite{tableformer, lysak_optimized_2023, dln}, quantitative evaluation of the overall document conversion accuracy on standard benchmark datasets will be subject to future work.  

\subsection{Benchmark Dataset}
\begin{figure}[t]
    \centering
    \includegraphics[width=0.9\linewidth]{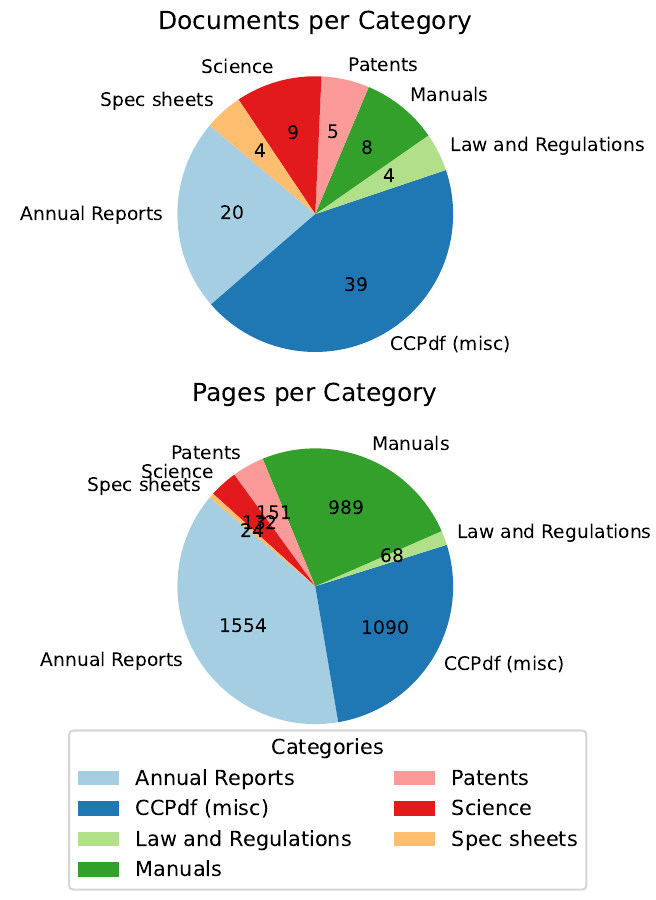}
    \caption{Dataset categories and sample counts for documents and pages.}
    \label{fig:dataset}
\end{figure}

To enable a meaningful benchmark, we composed a test set of 89 PDF files covering a large variety of styles, features, content, and length (see Figure~\ref{fig:dataset}). This dataset is based to a large extend on our DocLayNet~\cite{dln} dataset and augmented with additional samples from CCpdf~\cite{CCpdf} to increase the variety. Overall, it includes 4008 pages, 56246 text items, 1842 tables and 4676 pictures. As such, it is large enough to provide variety without requiring excessively long benchmarking times.
% As such, it is just large enough to provide variety while allowing benchmark experiments to complete in a time frame of at most several hours.

\begin{table*}[h]
\centering
\footnotesize
\caption{Versions and configuration options considered for each tested asset. \textsuperscript{*} denotes the default setting.}
\label{tab:experiments}
\begin{tabular}{rllll}
\toprule
\textbf{Asset}       & \textbf{Version} & \textbf{OCR}                       & \textbf{Layout}      & \textbf{Tables}                        \\ \midrule
Docling              & 2.5.2            & EasyOCR\textsuperscript{*}         & default              & TableFormer (fast)\textsuperscript{*}  \\ \midrule
Marker               & 0.3.10           & Surya\textsuperscript{*}           & default              & default                                \\ \midrule
MinerU               & 0.9.3            & auto\textsuperscript{*}            & doclayout\_yolo      & rapid\_table\textsuperscript{*}        \\ \midrule
Unstructured         & 0.16.5           & \multicolumn{3}{c}{\texttt{hi\_res} with table structure}                                   \\ \bottomrule
\end{tabular}

\end{table*}

\subsection{System Configurations\label{sec:sysconfig}}

We schedule our benchmark experiments each on two different systems to create reference numbers:
\begin{itemize}
    \item AWS EC2 VM (g6.xlarge), 8 virtual cores (AMD EPYC 7R13, x86), 32 GB RAM, Nvidia L4 GPU (24 GB VRAM), on Ubuntu 22.04 with Nvidia CUDA 12.4 drivers
    \item MacBook Pro M3 Max (ARM), 64GB RAM, on macOS 14.7
\end{itemize}

All experiments on the AWS EC2 VM are carried out once with GPU acceleration enabled and once purely on the x86 CPU, resulting in three total system configurations which we refer to as M3 Max SoC, L4 GPU, and x86 CPU.

% \subsection{Experiment setup}
\subsection{Benchmarking Methodology}

We implemented several measures to enable a fair and reproducible benchmark across all tested assets. Specifically, the experimental setup accounts for the following factors:
\begin{itemize}
    \item  All assets are installed in the latest available versions, in a clean Python environment, and configured to use the state-of-the-art processing options and models, where applicable. We selectively disabled non-essential functionalities to achieve a compatible feature-set across all compared libraries. % To harmonize the feature set, we disable additional features Docling does not support in the other assets, when possible.
    \item  When running experiments on CPU, we inform all assets of the desired CPU thread budget of 8 threads, via the \texttt{OMP\_NUM\_THREADS} environment variable and any accepted configuration options. The L4 GPU on our AWS EC2 VM is hidden. %, by setting \texttt{CUDA\_VISIBLE\_DEVICES=""} and installing CPU-only builds of AI runtimes such as pytorch, onnxruntime in the environment where applicable.
    \item When running experiments on the L4 GPU, we enable CUDA acceleration in all accepted configuration options, ensure the GPU is visible and all required runtimes for AI inference are installed with CUDA support.
\end{itemize}

Table~\ref{tab:experiments} provides an overview of the versions and configuration options we considered for each asset.

\subsection{Results}

\subsubsection{Runtime Characteristics}

\begin{figure}[htb]
    \centering
    \includegraphics[width=0.7\linewidth]{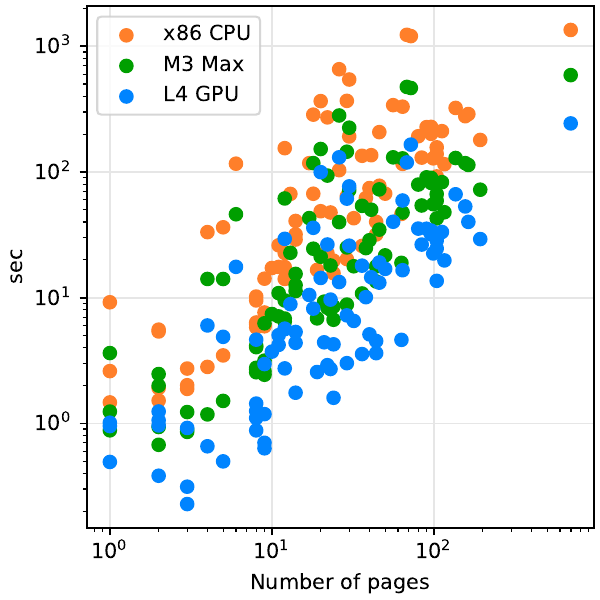}
    \caption{Distribution of conversion times for all documents, ordered by number of pages in a document, on all system configurations. Every dot represents one document. Log/log scale is used to even the spacing, since both number of pages and conversion times have long-tail distributions. }
    \label{fig:scatter}
\end{figure}

To analyze Docling's runtime characteristics, we begin by exploring the relationship between document length (in pages) and conversion time. As shown in Figure~\ref{fig:scatter}, this relationship is not strictly linear, as documents differ in their frequency of tables and bitmap elements (i.e., scanned content). This requires OCR or table structure recognition models to engage dynamically when layout analysis has detected such elements.

\begin{figure*}[!htb]
    \centering
    \includegraphics[width=0.8\linewidth]{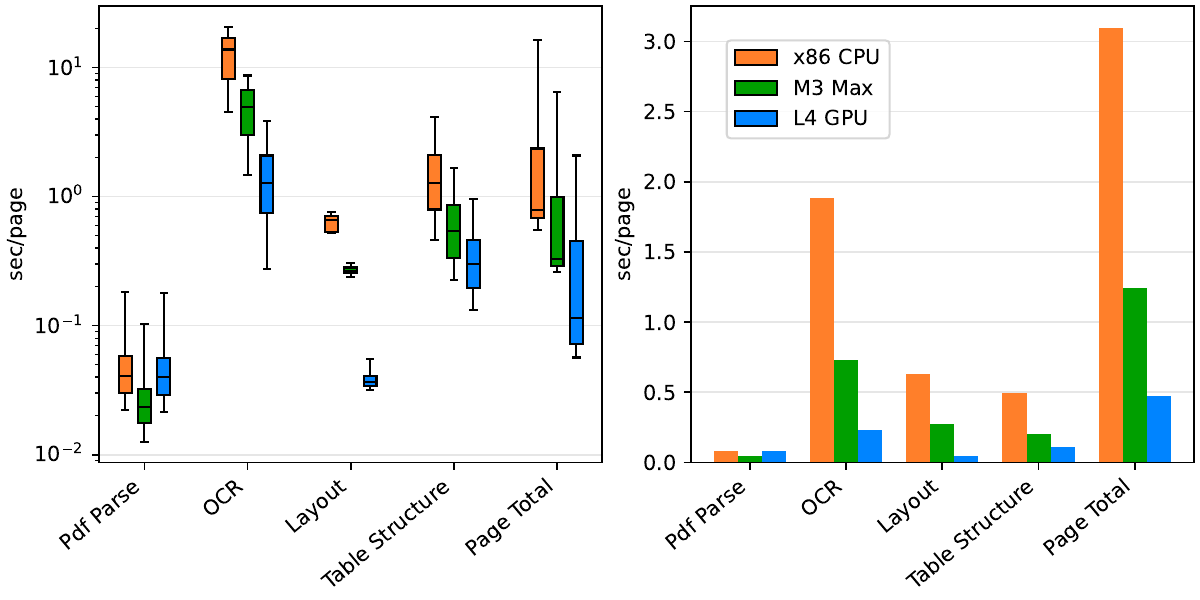}
    \caption{Contributions of PDF backend and AI models to the conversion time of a page (in seconds per page). Lower is better. Left: Ranges of time contributions for each model to pages it was applied on (i.e., OCR was applied only on pages with bitmaps, table structure was applied only on pages with tables). Right: Average time contribution to a page in the benchmark dataset (factoring in zero-time contribution for OCR and table structure models on pages without bitmaps or tables) .}
    \label{fig:docling_profiling}
\end{figure*}

By breaking down the runtimes to a page level, we receive a more intuitive measure for the conversion speed (see also Figure~\ref{fig:docling_profiling}). Processing a page in our benchmark dataset requires between 0.6 sec (5\textsuperscript{th} percentile) and 16.3 sec (95\textsuperscript{th} percentile), with a median of 0.79 sec on the x86 CPU. On the M3 Max SoC, it achieves 0.26/0.32/6.48 seconds per page (.05/median/.95), and on the Nvidia L4 GPU it achieves 57/114/2081 milliseconds per page (.05/median/.95). The large range between 5 and 95 percentiles results from the highly different complexity of content across pages (i.e., almost empty pages vs. full-page tables).   

Disabling OCR saves 60\% of runtime on the x86 CPU and the M3 Max SoC, and 50\% on the L4 GPU. Turning off table structure recognition saves 16\% of runtime on the x86 CPU and the M3 Max SoC, and 24\% on the L4 GPU. Disabling both OCR and table structure recognition saves around 75\% of runtime on all system configurations.

\subsubsection{Profiling Docling's AI Pipeline}

We analyzed the contributions of Docling's PDF backend and all AI models in the PDF pipeline to the total conversion time. The results are shown in Figure~\ref{fig:docling_profiling}. On average, processing a page took 481 ms on the L4 GPU, 3.1 s on the x86 CPU and 1.26 s on the M3 Max SoC.

% It is evident that applying OCR is the most expensive operation. In our benchmark dataset, OCR engages on 578 out of the 4008 total pages. On average, transcribing a page with EasyOCR took 1.6 sec on the L4 GPU, 13 sec on the x86 CPU and 5.1 sec on the M3 Max SoC. The layout model spent 44 ms on the L4 GPU, 631 ms on the x86 CPU and 270 ms on the M3 Max SoC on average for each page, making it the cheapest of the AI models, while TableFormer (fast flavour) spent 250 ms on the L4 GPU, 1.08 s on the x86 CPU and 436 ms on the M3 Max SoC on average per table, of which there are 1842. Regarding total time spent converting our benchmark dataset, TableFormer had less impact than other AI models, since tables appeared on only 28\% of all pages (see Figure~\ref{fig:docling_profiling}).

It is evident that applying OCR is the most expensive operation. In our benchmark dataset, OCR engages in 578 pages. On average, transcribing a page with EasyOCR took 1.6 s on the L4 GPU, 13 s on the x86 CPU and 5 s on the M3 Max SoC. The layout model spent 44 ms on the L4 GPU, 633 ms on the x86 CPU and 271 ms on the M3 Max SoC on average for each page, making it the cheapest of the AI models, while TableFormer (fast flavour) spent 400 ms on the L4 GPU, 1.74 s on the x86 CPU and 704 ms on the M3 Max SoC on average per table. Regarding the total time spent converting our benchmark dataset, TableFormer had less impact than other AI models, since tables appeared on only 28\% of all pages (see Figure~\ref{fig:docling_profiling}).

% On the L4 GPU, we observe a speedup of 8x (OCR), 14x (Layout model) and 4.3x (Table structure) compared to the x86 CPU and a speedup of 2.5x (OCR), 6x (Layout model) and 1.7x (Table structure) compared to the M3 Max CPU of our MacBook Pro. This shows that not all AI models equally benefit from GPU acceleration and there might be potential for optimization.

On the L4 GPU, we observe a speedup of 8x (OCR), 14x (Layout model) and 4.3x (Table structure) compared to the x86 CPU and a speedup of 3x (OCR), 6x (Layout model) and 1.7x (Table structure) compared to the M3 Max CPU of our MacBook Pro. This shows that there is no equal benefit for all AI models from the GPU acceleration and there might be potential for optimization.

% The time spent on parsing PDF through our docling-parse v2 backend is diminishingly small in comparison with the AI models. On average, parsing a PDF page took 82 ms on the x86 CPU and 44 ms on the M3 Max SoC (there is no GPU support).

The time spent in parsing a PDF page through our docling-parse backend is substantially lower in comparison to the AI models. On average, parsing a PDF page took 81 ms on the x86 CPU and 44 ms on the M3 Max SoC (there is no GPU support).

\subsubsection{Comparison to Other Tools}

\begin{figure}[!htb]
    \centering
    \includegraphics[width=0.8\linewidth]{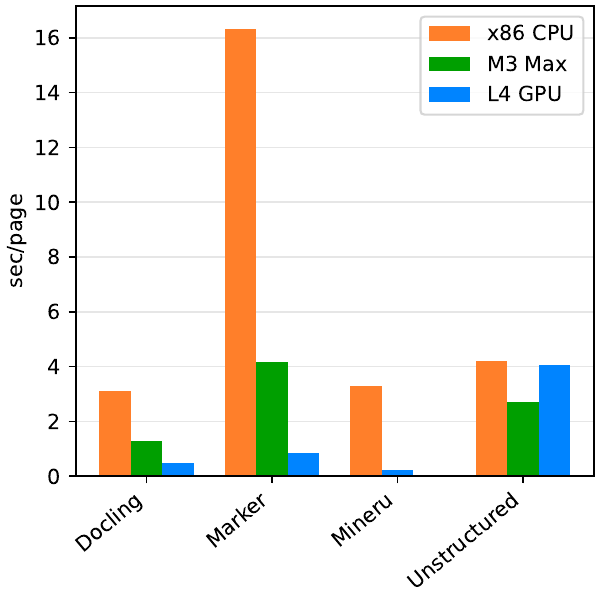}
    \caption{Conversion time in seconds per page on our dataset in three scenarios, across all assets and system configurations. Lower bars are better. The configuration includes OCR and table structure recognition (\texttt{fast} table option on Docling and MinerU, \texttt{hi\_res} in unstructured, as shown in table~\ref{tab:experiments}).}
    \label{fig:perf_comparison}
\end{figure}

We compare the average times to convert a page between Docling, Marker, MinerU, and Unstructured on the system configurations outlined in section~\ref{sec:sysconfig}. Results are shown in Figure~\ref{fig:perf_comparison}.

Without GPU support, Docling leads with 3.1 sec/page (x86 CPU) and 1.27 sec/page (M3 Max SoC), followed closely by MinerU (3.3 sec/page on x86 CPU) and Unstructured (4.2 sec/page on x86 CPU, 2.7 sec/page on M3 Max SoC), while Marker needs over 16 sec/page (x86 CPU) and 4.2 sec/page (M3 Mac SoC). MinerU, despite several efforts to configure its environment, did not finish any run on our MacBook Pro M3 Max. 
With CUDA acceleration on the Nvidia L4 GPU, the picture changes and MinerU takes the lead over the contenders with 0.21 sec/page, compared to 0.49 sec/page with Docling and 0.86 sec/page with Marker. Unstructured does not profit from GPU acceleration. 

%Additional experiments reveal that choosing the accurate version of TableFormer adds ~?? \% to the runtime (on all system configurations ??) over the fast version (default), while choosing the best-quality \texttt{structeq\_table}~\cite{xia2024docgenome} model on MinerU, a 1B parameter VLM for table structure recognition, results in a runtime of over 13 hours on the L4 GPU. 

\section{Applications}

Docling's document extraction capabilities make it naturally suitable for workflows like generative AI applications (e.g., RAG), data preparation for foundation model training, and fine-tuning, as well as information extraction.

As far as RAG is concerned, users can leverage existing Docling extensions for popular frameworks like LlamaIndex and then harness framework capabilities for RAG components like embedding models, vector stores, etc. These Docling extensions typically provide two modes of operation: one using a lossy export, e.g., to Markdown, and one using lossless serialization via JSON. The former provides a simple starting point, upon which any text-based chunking method may be applied (e.g., also drawing from the framework library), while the latter, which uses a swappable Docling chunker type, can be the more powerful one, as it can provide document-native RAG grounding via rich metadata such as the page number and the bounding box of the supporting context.
For usage outside of these frameworks, users can still employ Docling chunkers to accelerate and simplify the development of their custom pipelines.
%Last but not least, for advanced fine-grained cases, users can always access the \texttt{DoclingDocument} itself, however reusable chunking approaches can best be contributed back to the community as Docling chunkers.
Besides strict RAG pipelines for Q\&A, Docling can naturally be utilized in the context of broader agentic workflows for which it can provide document-based knowledge for agents to decide and act on.

% Moreover, document extraction can be an invaluable component to providing ground truth data. For instance, using Docling on textbooks and research papers can significantly contribute to domain-specific knowledge when infused to foundation model training and fine-tuning.

Moreover, Docling-enabled pipelines can generate ground truth data out of documents. Such domain-specific knowledge can make significant impact when infused to foundation model training and fine-tuning.

% Last but not least, Docling can be used as a backbone for information extraction tasks. Users who seek to create structured representations out of unstructured or semi-structured documents can leverage Docling for its streamlined pipeline, which  maps various document formats to the standardized, unified \texttt{DoclingDocument} format, as well as its strong table understanding capabilities that can help better analyze semi-structured document parts.

Last but not least, Docling can be used as a backbone for information extraction tasks. Users who seek to create structured representations out of unstructured documents can leverage Docling, which  maps various document formats to the unified \texttt{DoclingDocument} format, as well as its strong table understanding capabilities that can help better analyze semi-structured document parts.

\section{Ecosystem}

\begin{figure}[h]
    \centering
    \includegraphics[width=1\linewidth]{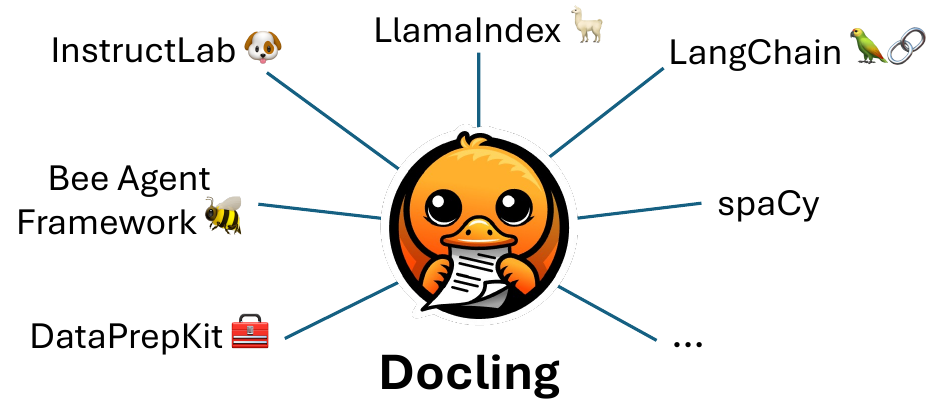}
    \caption{Ecosystem of Docling integrations contributed by the Docling team or the broader community. Docling is already used for RAG, model fine-tuning, large-scale datasets creation, information extraction and agentic workflows. }
    \label{fig:ecosystem}
\end{figure}

Docling is quickly evolving into a mainstream package for document conversion. The support for PDF, MS Office formats, Images, HTML, and more makes it a universal choice for downstream applications. % focused on enterprise document search, passage retrieval, LLM training and more.
Users appreciate the intuitiveness of the library, the high-quality, richly structured conversion output, as well as the permissive MIT license, and the possibility of running entirely locally on commodity hardware.

Among the integrations created by the Docling team and the growing community, a few are worth mentioning as depicted in Figure~\ref{fig:ecosystem}. For popular generative AI application patterns, we provide native integration within LangChain~\cite{Chase_LangChain_2022} and LlamaIndex~\cite{Liu_LlamaIndex_2022} for reading documents and chunking. Processing and transforming documents at scale for building large-scale multi-modal training datasets are enabled by the integration in the open IBM data-prep-kit~\cite{dpk}. Agentic workloads can leverage the integration with the Bee framework~\cite{bee}. For the fine-tuning of language models, Docling is integrated in InstructLab~\cite{instructlab}, where it supports the enhancement of the knowledge taxonomy.

Docling is also available and officially maintained as a system package in the Red Hat\textsuperscript{\tiny ®} Enterprise Linux\textsuperscript{\tiny ®} AI (RHEL AI) distribution, which seamlessly allows to develop, test, and run the Granite family of large language models for enterprise applications.

\section{Future Work and Contributions}

Docling's modular architecture allows an easy extension of the model library and pipelines. In the future, we plan to extend Docling with several additional models, such as a figure-classifier model, an equation-recognition model and a code-recognition model. This will help improve the quality of conversion for specific types of content, as well as augment extracted document metadata with additional information. Furthermore, we will focus on building an open-source quality evaluation framework for the tasks performed by Docling, such as layout analysis, table structure recognition, reading order detection, text transcription, etc. This will allow transparent quality comparisons based on publicly available benchmarks such as DP-Bench~\cite{zhong2020imagebasedtablerecognitiondata}, OmnidDocBench\cite{ouyang2024omnidocbenchbenchmarkingdiversepdf} and others. Results will be published in a future update of this technical report.

% The Docling roadmap is outlined in the discussions section~\footnote{\url{https://github.com/DS4SD/docling/discussions/categories/roadmap}} of our GitHub repository. \textbf{We encourage everyone to propose or implement additional features and models, and will gladly take your inputs and contributions under review}. The codebase of Docling is open for use and contribution, under the MIT license agreement and in alignment with our contributing guidelines included in the Docling repository. If you use Docling in your projects, please consider citing this technical report.

The codebase of Docling is open for use under the MIT license agreement and its roadmap is outlined in the discussions section~\footnote{\url{https://github.com/DS4SD/docling/discussions/categories/roadmap}} of our GitHub repository. We encourage everyone to propose improvements and make contributions.

\bibliography{aaai25}

\end{document}